# The AI Assessment Scale (AIAS): A Framework For Ethical Integration Of Generative AI In Educational Assessment

Mike Perkins [1*], Leon Furze [2], Jasper Roe [3], Jason MacVaugh [1]

[1] British University Vietnam, Vietnam.
[2] Deakin University, Australia
[3] James Cook University Singapore, Singapore.
[*] Corresponding Author: Mike.p@buv.edu.vn

## Abstract

Recent developments in Generative Artificial Intelligence (GenAI) have created a paradigm shift in multiple areas of society, and the use of these technologies is likely to become a defining feature of education in coming decades. GenAI offers transformative pedagogical opportunities, while simultaneously posing ethical and academic challenges. Against this backdrop, we outline a practical, simple, and sufficiently comprehensive tool to allow for the integration of GenAI tools into educational assessment: the AI Assessment Scale (AIAS).

The AIAS empowers educators to select the appropriate level of GenAI usage in assessments based on the learning outcomes they seek to address. The AIAS offers greater clarity and transparency for students and educators, provides a fair and equitable policy tool for institutions to work with, and offers a nuanced approach which embraces the opportunities of GenAI while recognising that there are instances where such tools may not be pedagogically appropriate or necessary.

By adopting a practical, flexible approach that can be implemented quickly, the AIAS can form a much-needed starting point to address the current uncertainty and anxiety regarding GenAI in education. As a secondary objective, we engage with the current literature and advocate for a refocused discourse on GenAI tools in education, one which foregrounds how technologies can help support and enhance teaching and learning, which contrasts with the current focus on GenAI as a facilitator of academic misconduct.

The AI Assessment Scale (AIAS): A Framework for Ethical Integration of Generative AI in Educational Assessment

# Introduction

Generative Artificial Intelligence (GenAI), a subfield of artificial intelligence, refers to models capable of generating data such as text, images, and audio, based on learned statistical patterns (Vaswani et al., 2017). The release of OpenAI's ChatGPT in November 2022 marked a turning point in the public adoption of GenAI, attracting over 100 million users in a matter of months (Milmo, 2023). GenAI applications developed by entities such as Microsoft and Google extend across multiple modes – text, visual, and audio – and are integrated into familiar educational platforms, highlighting the need to understand the implications of this technology in education. The growing prevalence of GenAI technologies in education presents an opportunity to rethink and potentially revolutionise existing pedagogical practices. GenAI is increasingly becoming a part of Higher Education (HE) discourse, offering new ways to teach, assess, and engage students. The adoption of this technology means that future learners will have access to radically new tools, as well as significant differences in their learning expectations.

While these technologies can bring about innovative changes in teaching and learning practices, they also raise important issues that educators, policymakers, and students must navigate. Adapting to these changing paradigms is not merely about incorporating new tools nor about banning them. It demands a nuanced understanding of how these technologies align with pedagogical objectives, and for educators and policymakers, academic integrity, ethical usage, and the development of critical thinking skills are of particular concern (Cotton et al., 2023; Mhlanga, 2023; Perkins et al., 2023; Rusandi et al., 2023). The impact of GenAI is not yet uniform across disciplines or educational models. While some fields have benefited from automation and data analytics, others, especially those requiring nuanced human judgment, have been less affected. Thus, the discourse surrounding GenAI in HE requires a multidimensional approach that accounts for ethical considerations, skill development, and engagement across diverse academic fields.

To date, the ethical and pedagogical implications of integrating GenAI into educational assessments remain underexplored in the academic literature. Although initial research (Smolansky et al., 2023) suggests that there may be gaps between students' and educators' opinions on how to achieve the best assessment approach, further research is required in this area. Similarly, while numerous studies have examined the technical capabilities of GenAI tools, few have explored the complexities of their ethical and effective integration across different educational models. An initial approach comes from Lim et al. (2023), who noted four 'paradoxes' of GenAI in HE that must be considered when planning for integration. To address these complexities, iterative, reflexive models that tackle GenAI issues yet allow for flexible development and fine-tuning must be developed.

This study presents an 'AI Assessment Scale' that provides clear directions and expectations to students regarding their engagement with GenAI tools for an assessed submission. It is designed to help educators consider how their assessments might need to be adjusted considering the prevalence of GenAI tools, clarify how and where GenAI tools might be used in their work, and support students in completing assessments in line with the principles of academic integrity. Although this scale has been designed for use in HE settings, it is sufficiently flexible for application in K-12 environments. In developing this scale, we consider the ethical challenges and considerations that emerge when GenAI is employed in HE assessments and provide a practical solution for higher education institutions (HEIs) to adapt to a GenAI assessment approach that enhances student engagement, ensures ethical usage, and fosters skill development.

# Literature

**Generative AI In Higher Education**

There are many positive applications of GenAI in HE, including its potential role in supporting students with understanding complex academic concepts and providing accommodations for students with communication disabilities (Kelly et al., 2023) or those learning in a second language. Furthermore, prompts for GenAI models, such as ChatGPT, can be tailored to specific educational purposes, such as summarising key concepts, generating exemplar material, supporting lesson planning, generating questions, and creating materials for simulations and role plays (Eager & Brunton, 2023). In the global HE landscape, Ansari et al's (2023) scoping review identified that almost one-third of studies related to the use of ChatGPT in HE focused on its use as a "teaching assistant", including for the creation of resources and lesson plans. The review also identifies ways in which students use the technology as a "personalised tutor" (p. 11) and the use of the tool by educators to assist in assessment and feedback. In specific disciplines such as computer science and software engineering, AI-powered code generation tools, such as Github's Copilot, have potential benefits for teaching programming (Bahroun et al., 2023). Consequently, there are potential affordances of GenAI technologies in HE which may balance the challenges.

However, whether the positives outweigh the negatives is not a settled debate. While Tilli et al. (2023) note a cautious optimism and positive tone in social media discourse surrounding ChatGPT in HE, and Fütterer et al.'s (2023) findings from an analysis of posts on *X* (Twitter) are similarly positive, other research has been conflicting. In a study of media headlines related to GenAI tools, Roe & Perkins (2023) found that the vast majority of headlines (including those on education) were concerned with the negative societal impacts of GenAI, and research emerging from late 2022 to early 2023 began to reflect significant concerns over academic dishonesty and the threat to traditional modes of assessment (Sullivan et al., 2023). Since then, while anxieties regarding academic integrity have not dissipated entirely (nor should they) discussions in HE and wider society have now started to become more nuanced and complex. The conversation has begun to move beyond simplistic comparisons of GenAI technologies to tools such as calculators and iPhones (Lodge, Yang, et al., 2023), and is shifting towards a more holistic understanding of how students and educators might use these technologies in more sophisticated ways. Lodge, Howard, et al. (2023), for example, propose the landscape of student use is much broader than simple information retrieval and automation, and may involve a complex network of self- and co-regulation with generative AI "agents".

Despite the evolving levels of nuance in our understanding of these technologies in HE, to date, there has been little discussion of the student perspectives of GenAI in HE in both the media and research (Sullivan et al., 2023).  In one of the few studies in this area, Kelly et al.'s (2023) survey of HE students suggests relatively low experience of the technologies and little confidence in their applications for learning and assessment. As with educators, students in HE may lack an understanding of the technical strengths and limitations of the technology, and those with less understanding may have higher levels of anxiety about the implications of the technology for their future studies and careers (Chan & Hu, 2023). This confirms the need for further exploration in not just how students are using GenAI, but how they perceive the effects of GenAI on their study.

# Problematizing The View Of GenAI Content As Academic Misconduct

## Redefining Academic Integrity in the Age of GenAI

Prior to the public release of GenAI tools, the early 2020s had already seen education stakeholders placing a renewed focus on academic misconduct and dishonesty, partly because of the COVID-19 pandemic, which led to perceived increases in cheating on behalf of students and teachers (Roe et al., 2023; Walsh et al., 2021). Simultaneously an 'arms race' (Cole & Kiss, 2000; Roe & Perkins, 2022) between technology-enabled academic misconduct and detection software (for example, automated paraphrasing tools) was already in full swing. In this broader context, the focus on academic integrity violations in the era of the GenAI tools can be seen as a one node in a network of existing conversations regarding the accelerating pace of digitalisation in HE and the resultant likelihood of what Dawson calls 'e-cheating', i.e. cheating that uses or is enabled by technology (Dawson, 2020, p. 4). However, conversations during the peak of COVID-19 centred more on the ability to complete assessed work with the help of external authors or contract cheating services, or to engage in traditional 'cheating' during remote examinations, while the present concern with GenAI tools as facilitators of misconduct tends to be focused on textual plagiarism and misrepresentation of authorship.

Taking a wider view, the issue at stake is the maintenance of integrity in the educational process: stakeholders want students to abide by the values of academic integrity. Although such values may vary slightly from institution to institution in terms of how they are encoded, they are best defined by the International Center for Academic Integrity as honesty, trust, fairness, respect, responsibility, and courage (ICAI, 2014). However, the discourse that AI-generated writing is by nature a violation of academic integrity and has no place in student-written work is, in our opinion, an unsustainable position for the future of HE.

There is now evidence from empirical research that major academic publishers of scholarly journals do not prohibit the use of GenAI; conversely, many encourage their use to refine and improve manuscripts, if their use is declared transparently and if the author takes full responsibility for the accuracy and veracity of the work (Perkins & Roe, 2023a). Pragmatically, this seems to be the only option that prepares academics and students for the rapid advances in AI; given that text detection services and combative approaches are flawed (Sadasivan et al., 2023), and a 'postplagiarism' world may be on the horizon (Eaton, 2023).

## Cultural Considerations

Aside from the difficulties of classifying when plagiarism may have occurred, a further consideration is the cultural values and belief systems of the user as to whether they are violating the norms of academic integrity. To date, no study has explored whether there is a cross-cultural difference in the perception of whether GenAI tools violate the norms of academic integrity. However, notions of ownership are strongly related to the Western individualist conception of knowledge production, and in a modern classroom setting, the way these norms are enacted in a codified system of rules and regulations does not reflect the diverse and varied attributes of student populations (Burke & Bristor, 2016). Put simply, it is a good moment to take a more inclusive approach that recognises the contextual, somewhat subjective view of plagiarism, particularly in the era of GenAI content, while keeping in mind the overarching goal of HE as a space for developing human-knowledge relationships (Kramm & McKenna, 2023). This view, in which AI is seen primarily as an additive and transformational tool rather than as an aid to misconduct, is what we see as a more helpful approach in considering the application of GenAI tools in HE.

**The Future of HE Assessment**

Educators need an alternative solution to support students in engaging with GenAI tools in an appropriate and ethical manner; one that also allows HEIs to maintain a standardised approach to dealing with the use of GenAI tools in an assessment situation. As a potential solution, we propose an AIAS in which educational institutions can adapt to their needs. The AIAS is a response to these broader concerns, amid calls to delineate the appropriate use of GenAI tools in education (Perkins & Roe, 2023b), design curricula with GenAI in mind (Bahroun et al., 2023), set clear guidelines for when and how GenAI can be used (Cotton et al., 2023), and support transparency in GenAI usage (Perkins & Roe, 2023a). Given that few global HEIs have developed clear policies for AI, let alone the more specific and novel field of Generative AI (Perkins & Roe, 2023b; Xiao et al., 2023), being able to employ a practical technique that fits within the wider constraints of a broader HE Institution policy is potentially of significant benefit to educators and students.

# The AI Assessment Scale

**Development and Rationale**

The development of the AIAS emerged from a shift in the perception and integration of GenAI tools in education. Initially, the use of GenAI tools in academic settings was largely viewed as a form of plagiarism or academic misconduct (Cotton et al., 2023; Perkins, 2023). Strict regulations were put in place prohibiting their use in student work, or assessment methodologies were switched to place greater emphasis on examinations or other assessment tasks which could be closely monitored (Fowler et al., 2023). However, as global understanding of and familiarity with these tools has expanded over a period of one year, a gradual shift in perception has occurred. The educational sector is now beginning to recognise the potential benefits of GenAI, albeit reluctantly, leading to a more nuanced approach to its integration.

This acceptance reflects a broader recognition that the use of technology in the workplace and in academic settings may have the ability to enhance performance and learning experiences.(Black & Lynch, 2001; Higgins et al., 2012; Oldham & Da Silva, 2015; Schacter, 1999). This change is part of a larger narrative where new technologies, initially met with scepticism and labelled as threats to genuine learning, gradually find their place as indispensable elements in education. Calculators, word processors, and the Internet have each shared their turn as 'the end of students doing real learning' only to become standard practice within a few years of general availability. Given that it is rare, at an organisational level, to turn down the opportunity to do more with fewer people at a higher level of performance (MacVaugh & Schiavone, 2010), the acceptance of technology in the workplace will, over time, trump the fear that HE is being degraded, even more so where programmes aim for authentic assessment with a deeper integration of GenAI tools to prepare students for professional life.

The AIAS has emerged as a response to these changing dynamics, highlighting the need for a more structured approach to the integration of GenAI in academic settings. In this context, the AIAS was conceptualised and developed. The initial concept, developed by the second author in discussions with the Education Faculty of Edith Cowan University, Perth, Australia in March 2023, was foreshadowed by media reports of the "Group of 8" universities in Australia reverting to pen and paper examinations to counter students using ChatGPT for assessments, thereby sparking a dialogue as to the potential of an alternative approach to GenAI tools and assessments.

The starting point for this development was to move away from a binary AI/no-AI approach, which began as a traffic light system, categorising the use of AI into 'No AI', 'Some AI', and 'Full AI'. This idea was refined over time to account for the more nuanced aspects of GenAI use in education to allow for some discretion on the part of both teachers and students, and to provide a scaffolded approach to assessment with GenAI. At lower levels of the scale, assessment tasks focus on ensuring students are able to obtain foundational knowledge and skills and develop a basic understanding of GenAI tools and their ethical considerations. At higher levels, the scale supports a deeper level of learner engagement with these technologies with more critical engagement and creative use expected.

This progressive approach requires faculty to consider how assessments may need to be restructured in light of GenAI tools and supports students' understanding of how GenAI tools can be effectively, and ethically used to support in their completion of assessments. The inherent challenge was to find a balance between simplicity (fewer scale points, easier for students and educators to understand) and clarity (more scale points, reducing ambiguity and allowing for alignment with institutional policies and individual educator requirements). Eventually, a five-point scale was developed, striking a balance between these two needs. However, we acknowledge that more or fewer scale points may be suitable for different institutions depending on their specific contexts and policies.

The AIAS is designed to achieve the following goals.

1. Help educators consider how their assessments might need to be adjusted in light of GenAI tools
2. Clarify to students how and where GenAI tools might be used in their work
3. Support students in completing assessments in line with the principles of academic integrity

The revised five-point AIAS is presented in Table 1.

The AI Assessment Scale (AIAS): A Framework for Ethical Integration of Generative AI in Educational Assessment## Scale Levels and Descriptions

| | | |
|---|---|---|
| 1 | **NO AI** | The assessment is completed entirely without AI assistance. This level ensures that students rely solely on their knowledge, understanding, and skills.<br><br>**AI must not be used at any point during the assessment.** |
| 2 | **AI-ASSISTED IDEA GENERATION AND STRUCTURING** | AI can be used in the assessment for brainstorming, creating structures, and generating ideas for improving work.<br><br>**No AI content is allowed in the final submission.** |
| 3 | **AI-ASSISTED EDITING** | AI can be used to make improvements to the clarity or quality of student created work to improve the final output, but no new content can be created using AI.<br><br>**AI can be used, but your original work with no AI content must be provided in an appendix.** |
| 4 | **AI TASK COMPLETION, HUMAN EVALUATION** | AI is used to complete certain elements of the task, with students providing discussion or commentary on the AI-generated content. This level requires critical engagement with AI generated content and evaluating its output.<br><br>**You will use AI to complete specified tasks in your assessment. Any AI created content must be cited.** |
| 5 | **FULL AI** | AI should be used as a "co-pilot" in order to meet the requirements of the assessment, allowing for a collaborative approach with AI and enhancing creativity.<br><br>**You may use AI throughout your assessment to support your own work and do not have to specify which content is AI generated.** |

*Table 1 The AI Assessment Scale*

## Introduction to the Scale

The AIAS provides a structured approach for HEIs to integrate GenAI into their assessment strategies, with each level specifying the extent of allowed GenAI use and student responsibility. The scale is flexible, recognising that while a linear model is beneficial for its simplicity, it must also accommodate the diverse nature of academic tasks and has been designed as a practical tool which we encourage HEIs to tailor based on their own policy decisions regarding the use of GenAI tools in academic evaluations.

Each level of the scale was intended to be cumulative, in terms of permitted AI engagement. For instance, a task designated at level 3 permits the use of GenAI tools for idea generation and structuring, as well as for editing the language. Level 4 is distinctive in that it requires students to not only use GenAI tools for specific tasks, but also to provide a critical evaluation of the AI's contribution. This level does not preclude the use of AI in the creative process but instead emphasises the need for student reflection and analysis of the AI-generated content. Choices as to whether students are allowed to use GenAI tools for editing the language of any such analysis would be up to the decision of the individual educator, depending on the requirements of each specific assessment task and broader institutional policy decisions.

**Supporting Guidelines for AI Use in assessment**

We recognise that providing a scale-based solution for GenAI tool usage needs additional context and urge HEIs to continue developing GenAI policies and student-facing guidelines that are flexible enough to cover the rapidly developing field while still allowing for the core elements of academic integrity to be considered. Recent work has demonstrated the slow speed of HEIs in creating formal policy documentation (Fowler et al., 2023; Perkins & Roe, 2023b; Xiao et al., 2023); however, guidelines and supporting multimedia content can be an effective way to provide additional context to how GenAI tools might be used in a safe and ethical manner. These guidelines may cover areas such as the ethics of GenAI tool usage, explaining how these tools can be cited and used in a transparent manner, exploring the limitations and biases of GenAI tools, and addressing security and privacy concerns.

For further examples and categorisations of assessment tasks across a range of disciplines, please refer to the supplementary material [available at this link](Perkins, Furze, et al., 2023).

**Scale Levels**

*Level 1: No AI*

At this level, students are not permitted to use GenAI in any form. This is appropriate for assessment tasks where it is preferable or necessary for students to rely solely on their own understanding, knowledge, or skills or where the use of GenAI is impractical or impossible. Although this stage may include the provision of technology-free examinations, it does not necessarily require examination conditions. For example:

- Technology-free discussions, debates, or other oral forms of assessment
- Technology-free ideation, individual, or group work in class
- Ad-hoc or planned viva-voce examinations, question and answer sessions, or formative discussions between students and educators

We recommend any Level 1 activities be conducted under supervision, or for low-stakes, formative assessments. This is due to potential equity concerns with permitting out-of-class work under "no AI" conditions, since students with English as a first language, higher levels of digital literacy, or access to better (often more expensive) GenAI tools may be able to use GenAI in ways which are potentially undetectable.

*Level 2: AI assisted idea generation and structuring*

At this scale level, students are permitted to use GenAI for brainstorming, gaining feedback, and structuring ideas; however, the final submission should not contain any content that was directly generated by AI. This level is useful for tasks in which students may benefit from extra support in developing ideas or improving their work, but in which the final product must be solely human-authored. Using GenAI tools at this level may benefit students by allowing them to explore a wider range of ideas and improve the depth or final quality of their work. Examples of Level 2 activities include:

- Collaborative brainstorming: Students can use AI to generate various ideas or solutions to a problem. These ideas can then be discussed, filtered, and refined by students in a collaborative setting.
- Structural outlines: Students may use AI to create a structured outline of their work.

The AI Assessment Scale (AIAS): A Framework for Ethical Integration of Generative AI in Educational Assessment- Research assistance: AI may be used to suggest topics, areas of interest, or even sources (using an Internet-connected model) that might be useful for a student's research.

### *Level 3: AI assisted editing*

At Level 3, students are permitted to employ generative AI for refining, editing, and enhancing the language or content of their original work. This may be particularly beneficial for non-native English speakers or those who struggle with language fluency. In a multimodal approach to assessment, GenAI tools might be permitted to support the editing of images or videos, but not for creating entirely new pieces. Examples include:

- Grammar, punctuation, and spelling: Students may use AI to identify and rectify grammatical, punctuation, spelling, and syntactical errors in their work.
- Word choice: AI can suggest appropriate or synonymous terms to replace simpler words and phrases, helping clarify writing.
- Structural edits: For students who may struggle to construct clear and coherent sentences, AI can assist in rephrasing for clarity without altering the original meaning.
- Visual editing: Image generation tools may be used to edit original images, such as through techniques like generative fill and generative expand (also referred to as in-painting and out-painting)

At this level, students are expected to submit their original work for comparison alongside AI-assisted content, thus ensuring the authenticity of their contributions. Assigning a Level 3 AI scale can make a traditional assessment task suitable for use in an AI-inclusive assessment environment, but it is more of a stop-gap approach which can be used until assessment tasks can be more fully adjusted to align with GenAI tool usage. Therefore, we recommend the use of this scale level as a transitional point in HEIs integration of GenAI tools.

### *Level 4: AI Task Completion, Human Evaluation*

At this level, students are requested or expected to use GenAI to complete specific portions of their tasks, but the emphasis remains on human evaluation and interpretation of the AI-generated content. Students must critically engage with and assess the AI outputs that they have created and evaluate their relevance, accuracy, and appropriateness. This level encourages a deeper understanding of the capabilities and limitations of GenAI tools, beyond basic text generation or editing. For example:

- Direct AI generation: Students may be tasked with using GenAI to produce content on a specific topic, theme, or prompt. This could range from generating datasets, social media posts, or crafting narratives. Students would use this as a basis for an original piece of work in which they may submit both the generated work and their own.
- Comparative analysis: After AI produces content, students may be asked to compare it with human-created content on the same topic, identifying differences, similarities, and areas of divergence. This can include comparisons with human-generated content.
- Critical evaluation: Students generate content with the express purpose of critiquing the output and questioning its choices, biases, and potential inaccuracies.
- Integration: Students may be tasked with integrating AI-generated content into a larger project to ensure cohesion and alignment with broader objectives. This might constitute part of an industry project or part of an authentic assessment task.

Level 4 introduces a more complex interplay between AI and student inputs. Here, students are expected to engage critically with AI outputs. This level is not prescriptive about the sequence in which AI and human intelligence interact; it allows for the possibility that students may or may not be allowed to use GenAI to aid in the rewriting process after conducting their analysis, but any GenAI content must be cited appropriately for transparency. This flexibility is intentional, acknowledging that the creative and iterative processes of academic work often do not follow a linear pathway. For example, they may conduct their own analysis and then refine or rework the output using GenAI tools. Deeper engagement with and evaluation of any GenAI-created content is an important element that defines Level 4 of the AIAS.

*Level 5: Full AI*

At the final level, AI may be used throughout the task at the student's discretion or teacher's suggestion. Assessments at this level may specify or recommend GenAI tools to be used, or may allow students' discretion in their choice.

Level 5 might be used in tasks which require the use of GenAI tools as part of addressing learning outcomes or when the skills and knowledge being assessed can be tested irrespective of AI usage. This level is also designed to allow for the exploration of GenAI as a collaborative and creative tool and reflects ways in which these technologies are being used outside of education, in fields such as journalism and marketing, where AI-generated content is increasingly used but still requires human editorial oversight (Hartmann et al., 2023; Kshetri et al., 2023). Example tasks include:

- Co-creation: Students are given broad themes or parameters in which they may achieve a task, and then actively iterate on GenAI content using a range of different tools and modes.
- GenAI exploration: Students use various GenAI tools to explore a wide range of ideas, styles, or solutions, exploring the ethical and practical implications of technology in a given domain.
- Real-time feedback loop: As students work on a task, they can continuously use GenAI to adjust their work, thereby shaping the final output.
- GenAI products: Students create finished products or artefacts using GenAI throughout, such as completed software or entire artworks.

**A Scalar Based Approach to GenAI Assessment**

The necessity for a scalar approach to integrating digital technologies in education becomes evident when considering the broad spectrum of digital tool usage, the imperative to delineate boundaries of academic honesty clearly, and the need for encouraging collaborative understandings between students and teachers regarding the pedagogical benefits of these technologies. Drawing on various studies, it becomes clear that the integration of GenAI and other digital tools into educational assessments and learning environments must move beyond simplistic, binary "use/don't use" approaches to accommodate the complex landscape of digital and information literacy, and ethical use.

Robinson and Glanzer (2017) demonstrate the importance of collaborative efforts among faculty and administrators in building a culture that encourages students to practice academic honesty. Through qualitative interviews with university students, the authors identified a need for collaborative construction of academic honesty guidelines and the unhelpfulness of simplistic and negative messages around academic integrity. This collaborative foundation is crucial for helping students understand the ethical use of digital technologies, including GenAI tools, in their academic work. Bretag et al. (2014) also support this argument, highlighting the need for Australian universities to adopt a holistic approach that goes beyond mere provision of information about academic integrity. The large survey of over 15,000 Australian HE students suggests universities need to do more to raise awareness of what does and does not constitute an academic integrity breach – one of the core purposes of the AIAS.

In the digital context, the confusion and challenges students face in understanding cyber-plagiarism are long-standing. Ercegovac (2005) underscores the need for clear guidelines and educational strategies that demystify the use of digital "objects" such as text and images and their fair use. This confusion is likely to be exacerbated in the context of GenAI, where intellectual property and copyright considerations are even more complex, and currently being argued in courts worldwide (e.g., Grynbaum & Ryan, 2023; Oremus & Izadi, 2024). Similarly, the work of Traphagan et al. (2014) emphasizes the necessity of integrating information literacy into courses to better prepare students for the ethical production and consumption of information in a digital age, including through the use of web-based platforms that may now involve GenAI. Even when encouraged to use digital tools, however, studies such as Bader et al. (2021) and Qayum and Smith (2015) reveal students often do not exploit the full pedagogical potential of the technology, instead focusing on their ease of use for information management, and basic information retrieval. These findings reveal a missed opportunity for deeper engagement with digital technologies that could enhance learning and assessment outcomes. The AIAS provides a structured framework that encourages exploration beyond basic uses – such as the use of GenAI for structural work in L2 or editing in L3 – and promotes an understanding of the broader pedagogical potential of GenAI and how students might ultimately use GenAI in critical and creative ways (L4), or for the entirety of a given task (L5).

**Balancing Skill Development, Engagement, and Ethics**

The AIAS helps support institutions and academic staff in balancing concerns related to academic integrity, student skill development, and meaningful engagement with both course content and assessment. Maintaining this balance during the current period of rapid transition towards the integration of GenAI tools in learning and teaching is a challenging task for educators, but one which we believe that the proposed AIAS can support with.

At its core, the AIAS is designed to promote the ethical use of GenAI tools by students, while fostering the development of both academic knowledge and, as we move to higher points on the scale, essential skills related to how these tools can be effectively used. By categorising the extent and manner of allowed GenAI usage in student work, the scale encourages students to critically evaluate how they incorporate AI into their assessment activities. Given that one of the objectives of the AIAS is to support students in avoiding academic misconduct by clearly defining the acceptable parameters of GenAI use, the scale helps shift the focus away from framing GenAI usage as a potential academic integrity concern to a focus on skill development and ethical engagement with the tools.

UNESCO (2023) have recognised the potential of GenAI tools to worsen 'digital poverty', with Bissessar (2023) also identifying the digital divide as an important consideration in the use of GenAI tools in classrooms. Access to GenAI tools can vary significantly among students, with some having access to more advanced or paid models than others do. This disparity raises concerns regarding equity in academic settings in the age of GenAI tools, especially in the Global South. If these tools are integrated into assessments, it is perhaps even more important to try and ensure equitable access, perhaps by standardising the GenAI tools permitted for use in assessments or limiting the use of certain advanced tools to maintain a level playing field.

However, as new GenAI tools continue to emerge, addressing this in a practical manner has become a significant challenge.

The primary purpose of the AIAS is designed as a tool to assist students in contextualising how GenAI tools should be integrated into any assessment task. Engaging students in a clear discussion of where AI is and is not appropriate for each task may contribute to students harnessing these technologies effectively and ethically, aligning with the broader trend of GenAI acceptance in HE. Although this scale may also be used to support academics in making a judgement as to whether GenAI tools have been misused by students, we would caution against framing this misuse as a potential breach of academic integrity and instead suggest that any grade penalties for students misusing these tools can be resolved through the application of assessment rubrics which set boundaries for the use of technological aids supporting cognitive offloading (Dawson, 2020; Risko & Gilbert, 2016). While there is a need for institutional rules around the allowable use of GenAI tools to ensure a high standard of academic integrity, which may incorporate sanctions or punishments, this should not be the focus of the development of effective policy decisions related to the integration of GenAI into assessment, teaching, and learning in HE.

# Conclusion

**Limitations and Future Research**

We propose the AIAS as a tool which can support both educators and students in coming to a shared understanding of the acceptable use of GenAI tools in an assessment, and that work is submitted by students with a fundamental focus on the values of responsibility and accuracy. However, we recognise that these scales are constrained by limitations and may not be suitable for all forms of assessments. Given the broad range of subjects and materials taught across K-12 and HE, a scalar approach must be modified and customised in line with programme or module learning outcomes. Furthermore, while offering a scale may reduce both unintentional and intentional acts of academic misconduct, this approach does not eliminate the opportunity for students to try and gain an unfair advantage over others–those who wish to could still claim to be submitting original, independent work while disguising their use of GenAI tools.

As a proposed tool which has only recently been put into operation in a pilot cohort, there is also no empirical research on the scale's effectiveness in combating academic misconduct or improving student outcomes. A further limitation of this model is the rapidly advancing nature of AI technologies, meaning that any scale will need to be continuously maintained and adapted as new GenAI tools become commonplace, and approaches to assessment move towards being more AI inclusive.

Although the research base in GenAI and education is growing rapidly, there is still a dire need for additional empirical work in this area. Research is needed to understand how and why students use GenAI tools in HE for learning and assessment and how responsible use can be encouraged. As part of this effort, empirical studies on the use of the AIAS in diverse educational settings can contribute. It is important that research also takes a long-term view and considers the viability of traditional forms of assessment in the face of a growing number of paradigm-shifting technologies.

**Summary**

The discourse surrounding GenAI tools in education, particularly since the advent of ChatGPT in November 2023, has undergone a rapid transformation. Initially, the focus was on curbing the use of these tools through policy adjustments and alterations in assessment strategies, and the emergence of AI text detection tools further bolstered this approach, as HEIs saw them as instrumental in identifying the misuses of this technology. However, recent studies have highlighted significant limitations of these detection tools, including challenges with accuracy, risks of false accusations, and potential biases against non-native speakers.

These findings have spurred growing recognition among educational institutions of the need to reconsider reliance on detection tools. There is an increasing call for an alternative approach in which academic integrity remains paramount, but the use of GenAI tools is integrated to foster student skill development, particularly in preparation for future workplaces where these tools might be prevalent. The AIAS has emerged as a vital tool in this context. It reframes the conversation with students about GenAI from a prohibitive stance to a more constructive one, guiding them on how to use these tools effectively, using a five-point scale designed to support a balance between simplicity and clarity.

Nonetheless, challenges remain, notably regarding issues of accessing GenAI software, practical applications in diverse settings, and managing the digital divide. Access to GenAI tools may vary by location, and premium versions of some common tools, such as GPT-4, command monthly payments. Although tools such as the AIAS demonstrate one way in which HEIs can effectively integrate GenAI tools to ensure continued student engagement and skill development, this needs to be balanced with the broader institutional requirements of maintaining academic integrity, changing expectations of learners, and future requirements of industry. This requires a considerable amount of institutional agility and open dialogue with students to better equip new generations to harness GenAI tools as part of their future learning and professional pathways.

Despite these challenges, it is important to recognise the value of actively discussing GenAI with students. By doing so, educators not only prepare them for a future increasingly influenced by these technologies but also promote an environment of informed and responsible use. As GenAI continues to evolve and become an integral part of professional and academic landscapes, equipping students with the knowledge and skills necessary to navigate these changes is crucial. The AIAS is a step towards achieving this goal, fostering an educational environment that is adaptive, ethical, and forward-thinking.

## Conflict of Interest

The author(s) disclose that they have no actual or perceived conflicts of interest. The authors disclose that they have not received any funding for this manuscript beyond resourcing for academic time at their respective university.

This study used Generative AI tools to produce draft text, and revise wording throughout the production of the manuscript. Multiple modes of ChatGPT over different time periods were used, with all modes using the underlying GPT-4 Large Language Model. The authors reviewed, edited, and take responsibility for all outputs of the tools used in this study.

The initial version of this manuscript prior to peer-review was posted on the arXiv preprint server andis available undera Creative Commons Attribution-NonCommercial-NoDerivatives4.0 International (CC BY-NC-ND 4.0) licence at https://arxiv.org/abs/2312.07086 .